\documentclass[conference]{IEEEtran}
\IEEEoverridecommandlockouts

\usepackage{cite}
\usepackage{amsmath,amssymb,amsfonts}
\usepackage{textcomp}
\usepackage{xcolor}

\usepackage{booktabs}  
\usepackage[hidelinks, bookmarks=false]{hyperref}  
\usepackage{mathtools}

\usepackage{algorithm}
\usepackage{algpseudocode}

\usepackage{multirow}  
\usepackage{capt-of}  
\usepackage[defaultlines=2,all]{nowidow}  
\usepackage{url}  
\usepackage{pgfplots}
\usepackage{tikz}
\usepackage{siunitx}

\usepackage{caption}     
\usepackage{subcaption}

\def\BibTeX{{\rm B\kern-.05em{\sc i\kern-.025em b}\kern-.08em
    T\kern-.1667em\lower.7ex\hbox{E}\kern-.125emX}}

\begin{document}

\title{
MUFFLe: Efficient Model Update Compression via Generalized Deduplication for Federated Learning
\thanks{This work has been supported in part by the Forever Bearing project (Grant No. 4353-00017B) and the CRISPER-IoT project (Grant No. 5364-00007B) granted by the  Innovation Foundation Denmark.}
}

\author{
\IEEEauthorblockN{Xiaobo Zhao, Daniel E. Lucani}
\IEEEauthorblockA{
DIGIT, Department of Electrical and Computer Engineering,
Aarhus University\\
\{xiaobo.zhao, daniel.lucani\}@ece.au.dk}
}

\maketitle

\begin{abstract}
Federated learning is well suited to edge environments but is often limited by the uplink cost of transmitting model updates.
This Work-in-Progress paper presents MUFFLe, a communication-efficient update compression scheme that integrates generalized deduplication (GD) into the FedAvg pipeline.
MUFFLe deduplicates repeated patterns across the update vector, yielding a fixed-rate, variable-count compression scheme.
Preliminary experiments on IID MNIST with 20 clients show that MUFFLe reaches the target accuracy of $92.93\%$ with 38~MB cumulative uplink communication, compared with 75~MB for 8-bit quantization, 86~MB for Top-$k$ sparsification, and 310~MB for uncompressed FedAvg.
These results demonstrate the feasibility of applying GD to communication-efficient federated learning.
\end{abstract}

\section{Introduction}

Federated learning (FL) enables distributed model training across edge devices without centralizing raw data~\cite{mcmahan2017communication}.
In practical edge deployments, however, client-to-server update transmission often becomes the main bottleneck due to limited uplink bandwidth, non-negligible latency, and increased device energy consumption~\cite{lim2020federated,vahabi2025federated}.
A common way to reduce communication cost is to compress local updates before transmission.
Representative approaches include quantization~\cite{reisizadeh2020fedpaq}, which reduces the bit-width of each value, and Top-$k$ sparsification~\cite{stich2018sparsified}, which transmits a subset of update entries.

\begin{figure}[t]
\centering
\begin{subfigure}[c]{0.35\columnwidth}
	\centering
	\includegraphics[width=\linewidth]{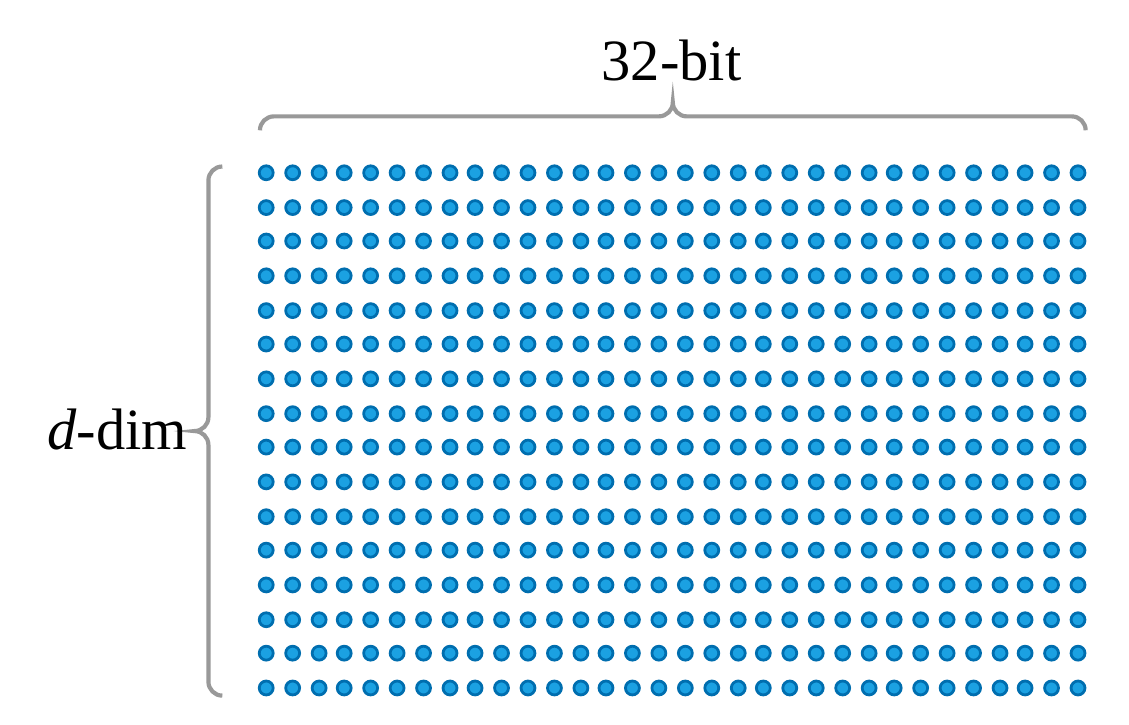}
	\caption{Uncompressed}
	\label{fig:intro_uncompressed}
\end{subfigure}
\begin{subfigure}[c]{0.35\columnwidth}
	\centering
	\includegraphics[width=\linewidth]{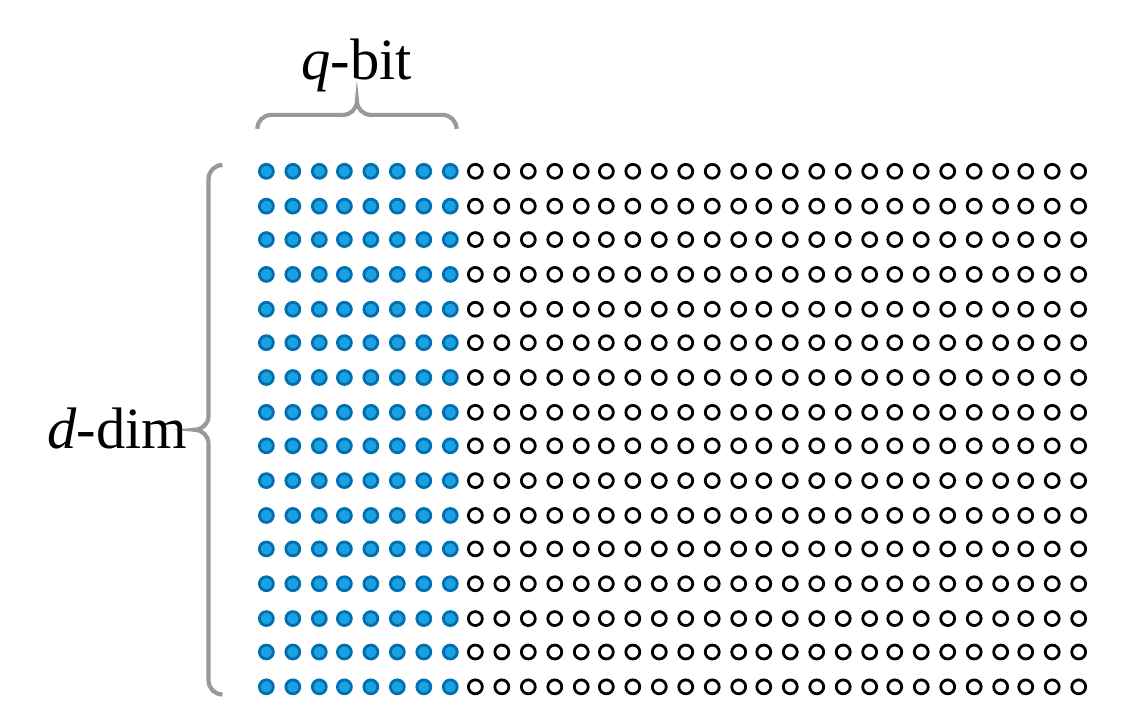}
	\caption{Quantization}
	\label{fig:intro_quant}
\end{subfigure}

\vspace{0.5em}

\begin{subfigure}[c]{0.35\columnwidth}
	\centering
	\includegraphics[width=\linewidth]{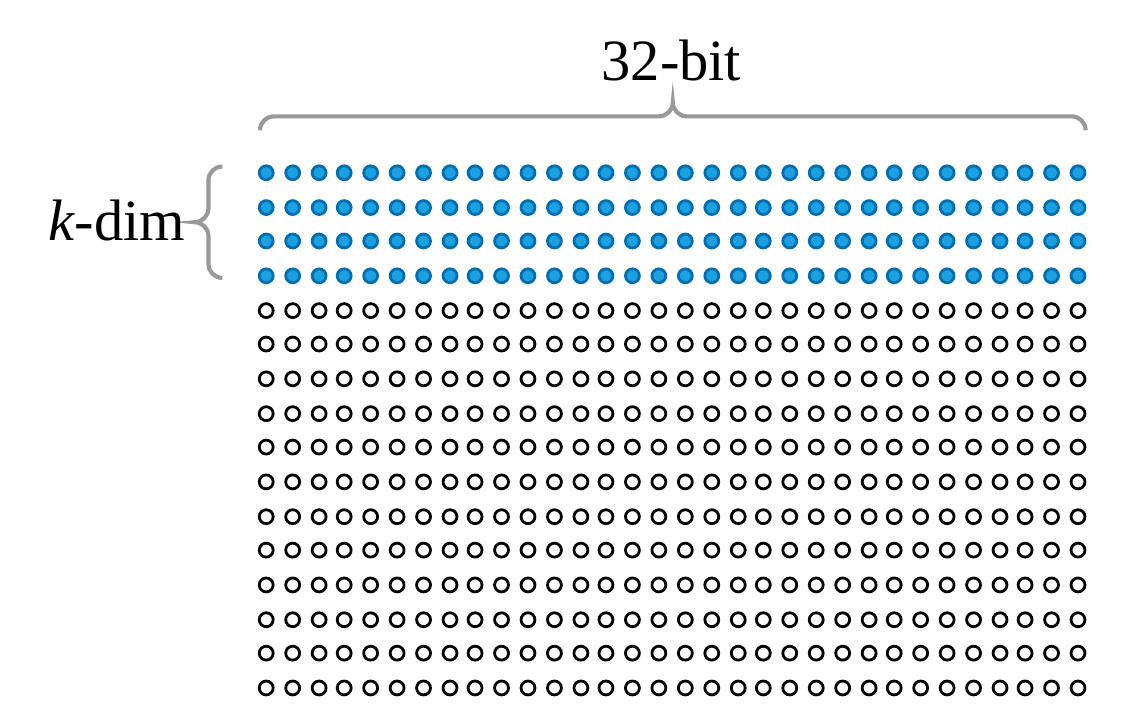}
	\caption{Top-k}
	\label{fig:intro_topk}
\end{subfigure}
\begin{subfigure}[c]{0.35\columnwidth}
	\centering
	\includegraphics[width=\linewidth]{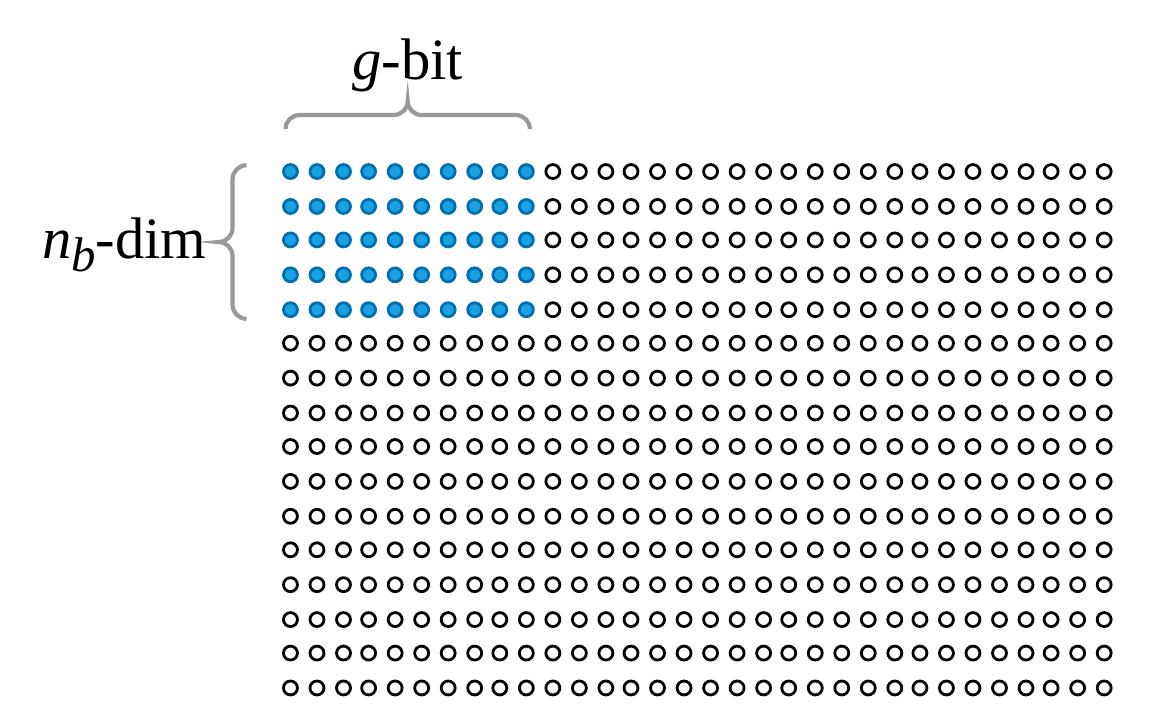}
	\caption{MUFFLe}
	\label{fig:intro_gdfl}
\end{subfigure}
\caption{Comparison of update-transmission strategies considered in this work: (a) uncompressed communication, (b) Quantization, (c) Top-k sparsification, and (d) MUFFLe.
}
\vspace{-5mm}
\label{fig:intro_comparison}
\end{figure}

In this paper, we investigate a different direction based on generalized deduplication (GD)~\cite{vestergaard2019_a, Vestergaard_2019b,Vestergaard_2020, zhao2026entrogd}.
Specifically, MUFFLe adapts GD to lossy federated update compression by retaining non-constant most-significant bits and deduplicating repeated base patterns across update entries.
This yields a fixed-rate, variable-count compression scheme, where the retained bit-width is fixed while the number of transmitted bases depends on the redundancy structure of the update.
Although MUFFLe may appear similar to a combination of quantization and Top-$k$ sparsification, as shown in Fig.~\ref{fig:intro_comparison}, it is fundamentally different in that it reduces communication by exploiting repeated patterns across the full update, rather than by uniformly lowering precision or transmitting only a subset of entries.
Moreover, MUFFLe is orthogonal to these strategies and can be combined with quantization or sparsification in future work.
We present this study as a proof of concept in a preliminary Work-in-Progress (WiP) setting.
Experiments on independent and identically distributed (IID) MNIST~\cite{lecun2002gradient} with 20 clients and a lightweight multilayer perceptron (MLP) under FedAvg~\cite{mcmahan2017communication} show that MUFFLe achieves the target accuracy with substantially lower cumulative uplink communication than uncompressed FedAvg, quantization, and Top-$k$ sparsification.

The main contribution of this paper is the introduction of MUFFLe as a communication-efficient compression scheme for federated learning update transmission.
We also provide a preliminary empirical evaluation against standard compression baselines under a common communication-oriented metric.

\section{Problem Formulation}

\subsection{The Federated Averaging (FedAvg) Algorithm}
As the primary algorithmic framework for FL, FedAvg trains a global model $w \in \mathbb{R}^d$ across $U$ edge clients by solving:
\vspace{-3mm}
\begin{equation}
    \min_{w \in \mathbb{R}^d} F(w) = \sum_{u=1}^U \frac{n_u}{N_{tot}} F_u(w),
\end{equation}
where $n_u$ is the number of samples at client $u$, $N_{tot}$ is the total dataset size, and $F_u(w)$ is the local loss.

At communication round $t$, the server selects clients $\mathcal{S}_t$ ($|\mathcal{S}_t| = m$) and broadcasts $w_t$.
Each selected client computes:
\begin{equation}
    \Delta w_t^u = w_{t,E}^u - w_t,
\end{equation}
where $w_{t,E}^u$ is the local model state after $E$ epochs of SGD.
The server then performs a weighted model aggregation:
\begin{equation}
    w_{t+1} = w_t + \sum_{u \in \mathcal{S}_t} \frac{n_u}{N_{\mathcal{S}_t}} \Delta w_t^u,
\end{equation}
where $N_{\mathcal{S}_t} = \sum_{u \in \mathcal{S}_t} n_u$ is the total number of samples across the selected clients.

\subsection{Communication-Efficient Formulation}
In resource-constrained edge networks, transmitting full-precision updates $\Delta w_t^u$ incurs substantial latency and energy overhead.
We use a compression operator $\mathcal{C}: \mathbb{R}^d \to \mathcal{M}$ to map each local update to a compact message, and a reconstruction mapping $\mathcal{D}: \mathcal{M} \to \mathbb{R}^d$ to approximate it at the server:
\begin{equation}
    \hat{\Delta w}_t^u = \mathcal{D}(\mathcal{C}(\Delta w_t^u)).
\end{equation}
The objective is to minimize the cumulative communication volume $\mathcal{B}$ needed to reach target accuracy $\mathcal{A}_{target}$:
\begin{equation}
    \min_{\mathcal{C}} \mathcal{B} = \sum_{t=1}^{T^*} \sum_{u \in \mathcal{S}_t} \text{size}\left( \mathcal{C}(\Delta w_t^u) \right),
\end{equation}
where $\text{size}(\cdot)$ is the message bit-length, and $T^*$ is the first round at which the global model meets the performance threshold. This captures the communication-efficient FL trade-off: aggressive compression lowers per-round latency and energy but can introduce reconstruction errors that increase the convergence rounds $T^*$.

\subsection{Baseline Compression Schemes}
We benchmark two standard compression schemes.

\textbf{Quantization:} This scheme maps each scalar in $\Delta w_t^u$ to a $q$-bit signed integer. Given $\alpha = \|\Delta w_t^u\|_\infty$, the scale factor is $S = \alpha / (2^{q-1}-1)$. The encoding mapping is $\mathcal{C}_Q(x) = \text{round}(x/S)$, and the server reconstructs the update with $\mathcal{D}_Q(i) = i \cdot S$. The transmission cost is $(d \times q) + 32$ bits: $d$ quantized parameters plus one 32-bit float for $S$.
    
\textbf{Top-$k$:} This scheme keeps the $k$ elements with the largest absolute magnitudes. The client transmits $k$ tuples, each containing a 32-bit floating-point value $v_j$ and index $j \in \{1, \dots, d\}$.
The reconstruction mapping $\mathcal{D}_k(\cdot)$ is given by:
    \begin{equation}
        [\hat{\Delta w}_t^u]_j = 
        \begin{cases} 
        v_j & \text{if } j \in \text{transmitted indices} \\
        0 & \text{otherwise}
        \end{cases}
    \end{equation}
    The compressed bit-length is $k(32 + \lceil \log_2 d \rceil)$, where $\lceil \log_2 d \rceil$ bits encode each index.

\section{Methodology: MUFFLe}

We propose \emph{MUFFLe}, a communication-efficient compression scheme for client-to-server model updates in federated learning. Let $\Delta w_t^u \in \mathbb{R}^d$ denote the local update generated by client $u$ at communication round $t$. In contrast to conventional quantization, which assigns a fixed bit-width to every element, and Top-$k$ sparsification, which transmits a number of selected elements, MUFFLe combines fixed-rate symbol truncation with redundancy-aware deduplication across the update vector.

The key idea is to convert the update into a sequence of fixed-length binary symbols and exploit repeated patterns among their most-significant bits. This yields a \emph{fixed-rate, variable-count} compression scheme: the bit-width per symbol is fixed, while the number of distinct transmitted symbols depends on the redundancy structure of the update.

\begin{figure}[t]
\centering
\includegraphics[width=0.9\columnwidth]{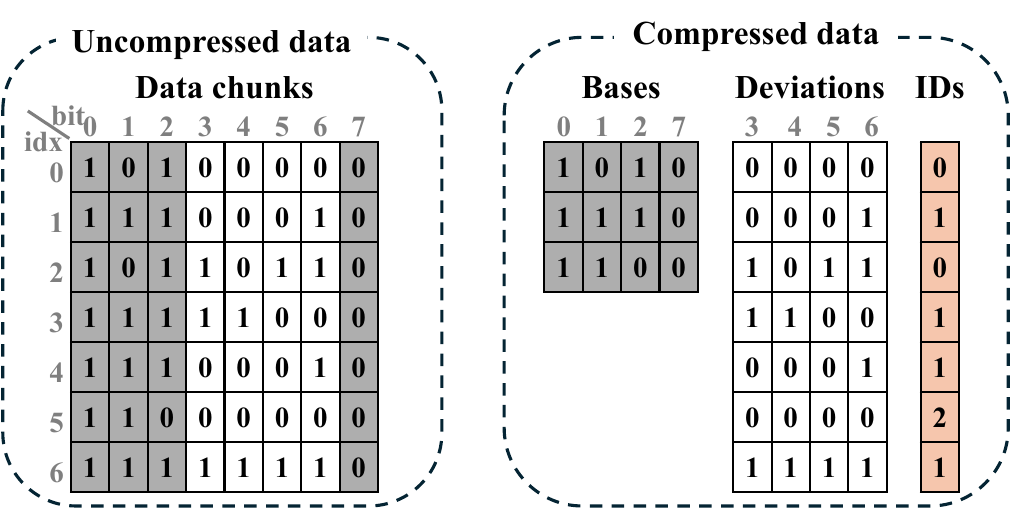}
\caption{An example of GD applied to 8-
bit data chunks.}
\label{fig:gd_example}
\end{figure}

\subsection{Generalized Deduplication (GD)}

As illustrated in Fig.~\ref{fig:gd_example}, Generalized Deduplication (GD) is a compression framework that partitions data chunks into frequently recurring bit patterns, termed \textit{bases}, and high-variance residual components, termed \textit{deviations}.
It achieves lossless compression by deduplicating the bases and storing the deviations together with the corresponding base IDs.
For example, chunks 0 and 2 in Fig.~\ref{fig:gd_example} share the common base \texttt{101....0}, which is stored only once and referenced by base ID $0$, while their distinct deviations are recorded separately.
Compression is effective when the number of unique bases is substantially smaller than the total number of data chunks.

\subsection{MUFFLe}
MUFFLe adapts GD from lossless data compression to lossy client-update compression in federated learning.
Since floating-point data can be compressed more effectively after being scaled and transformed into an unsigned integer representation~\cite{hurst2024greedygd}, we follow the preprocessing steps in~\cite{hurst2024greedygd} to convert the local update $\Delta w_t^u$ into a binary integer representation.
MUFFLe first identifies the constant bits shared across all $d$ update elements and retains only a predefined number of $g$ non-constant most-significant bits (MSBs) for each element.
These retained MSBs form the base representation of each update element and establish the fixed bit-rate of the scheme.

Deduplication is then performed over the truncated MSB bases.
Specifically, update elements with the same retained $g$-bit MSB pattern are mapped to a single base entry and represented by the corresponding base ID.
Unlike existing GD-based methods, which store deviations to enable lossless reconstruction, MUFFLe omits them to obtain a lossy approximation of the update.
As illustrated in Fig.~\ref{fig:intro_gdfl}, MUFFLe therefore reduces uplink communication through two complementary mechanisms: fixed-rate truncation decreases the number of bits retained per element, while deduplication avoids transmitting replication across the updates.

In this work-in-progress study, we do not apply error feedback or compensation to the considered compression methods.
A future extension could store the discarded deviation bits as residual information and accumulate them into the next round's update.
Since these deviations require fewer bits than residuals stored in the original high-precision floating-point format, this extension may also reduce device RAM usage.

The transmitted representation therefore consists of a dictionary of unique bases and a sequence of base IDs, one for each update element.
Since the redundancy structure of the update can vary across communication rounds, the number of unique bases is data-dependent.
The total compressed cost of MUFFLe is $n_b \cdot g + d \cdot \lceil \log_2(n_b) \rceil$, where $n_b$ denotes the number of unique base symbols transmitted, $g$ denotes the non-constant base bit length, and $\lceil \log_2(n_b) \rceil$ denotes the number of bits required to encode the base ID of each update element.
Since the constant bits are shared across all update entries and stored only once, and metadata overhead is negligible, they are omitted from the total bit calculation.
Upon receiving the compressed stream, the server reconstructs the update $\hat{\Delta w}_t^u$ by mapping the received symbols to their original indices according to the IDs and then converting the binary representation back into the approximate update vector.

\section{Experimental Results}
We evaluate the communication efficiency of MUFFLe on MNIST under an IID partition, with the training set uniformly distributed across $K=20$ clients.
In each communication round, a fraction $C=0.2$ of clients is selected uniformly at random, yielding $n=4$ participating clients.
The model is a lightweight MLP with a flattening layer, a 128-unit ReLU hidden layer, and a 10-class Softmax output layer.
Local training uses SGD with learning rate $\eta=0.01$ and $E=1$ local epoch.
Fixed random seeds ensure reproducible weight initialization and client selection.

All methods compress client-to-server parameter updates.
We compare MUFFLe with:
(i) \textbf{Uncompressed}: 32-bit FedAvg;
(ii) \textbf{8-bit Quantization}; and
(iii) \textbf{Top-$k$ Sparsification}, with $k$ matched to the 8-bit quantization budget.
For simplicity, updates are rounded to four decimal places before MUFFLe preprocessing.
Communication efficiency is measured by \textit{cumulative bits}, defined as the total uplink transmission across all participating clients.
We report the communication cost required to reach the target accuracy, defined as the test accuracy of the uncompressed baseline at 200 rounds.

\begin{figure}[ht]
\centering
\begin{subfigure}[b]{0.49\columnwidth}
	\centering
	\includegraphics[width=\linewidth]{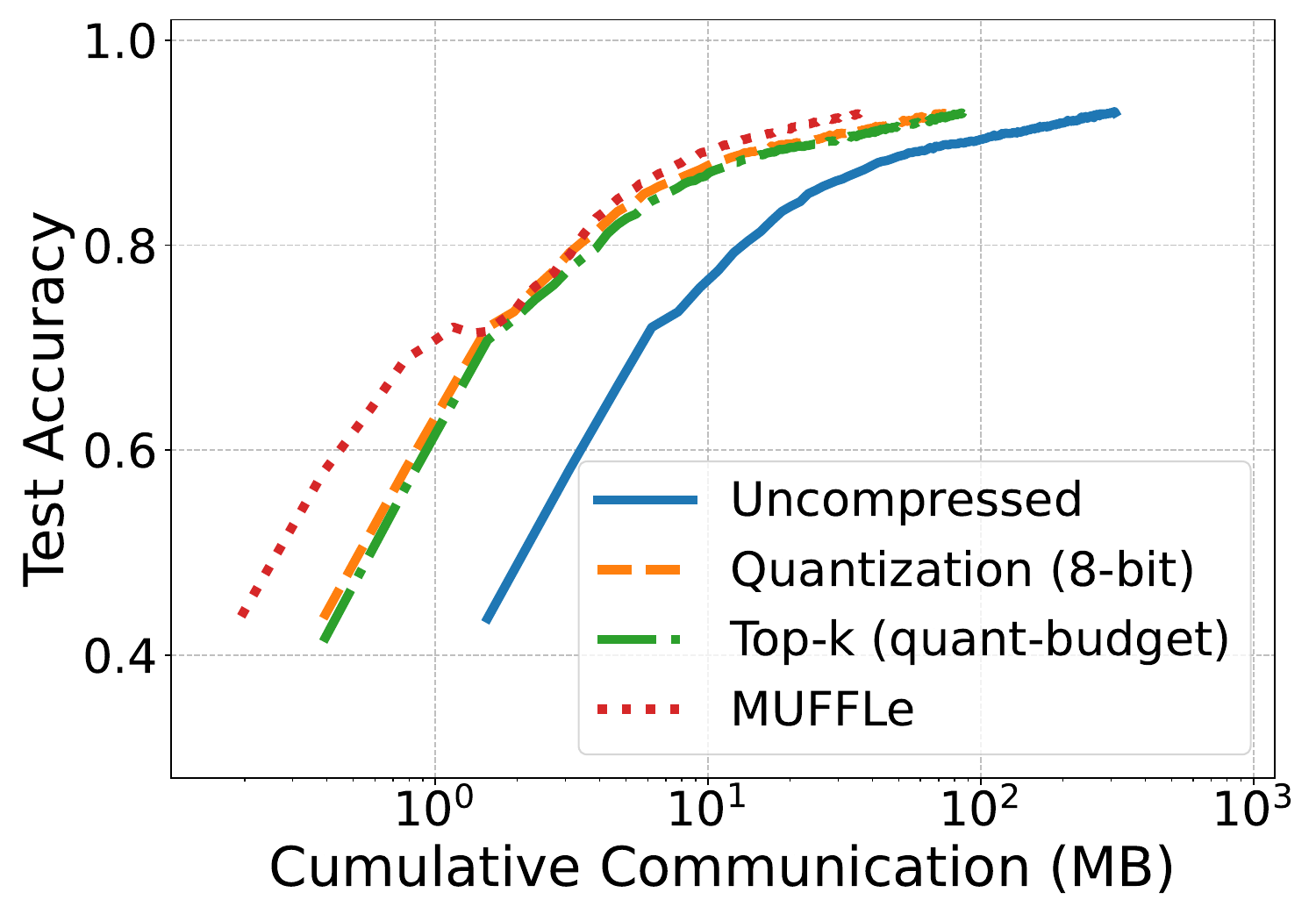}
	\caption{Acc. vs comm. cost.}
	\label{fig:acc_vs_comm_cost}
\end{subfigure}
\hfill
\begin{subfigure}[b]{0.49\columnwidth}
	\centering
	\includegraphics[width=\linewidth]{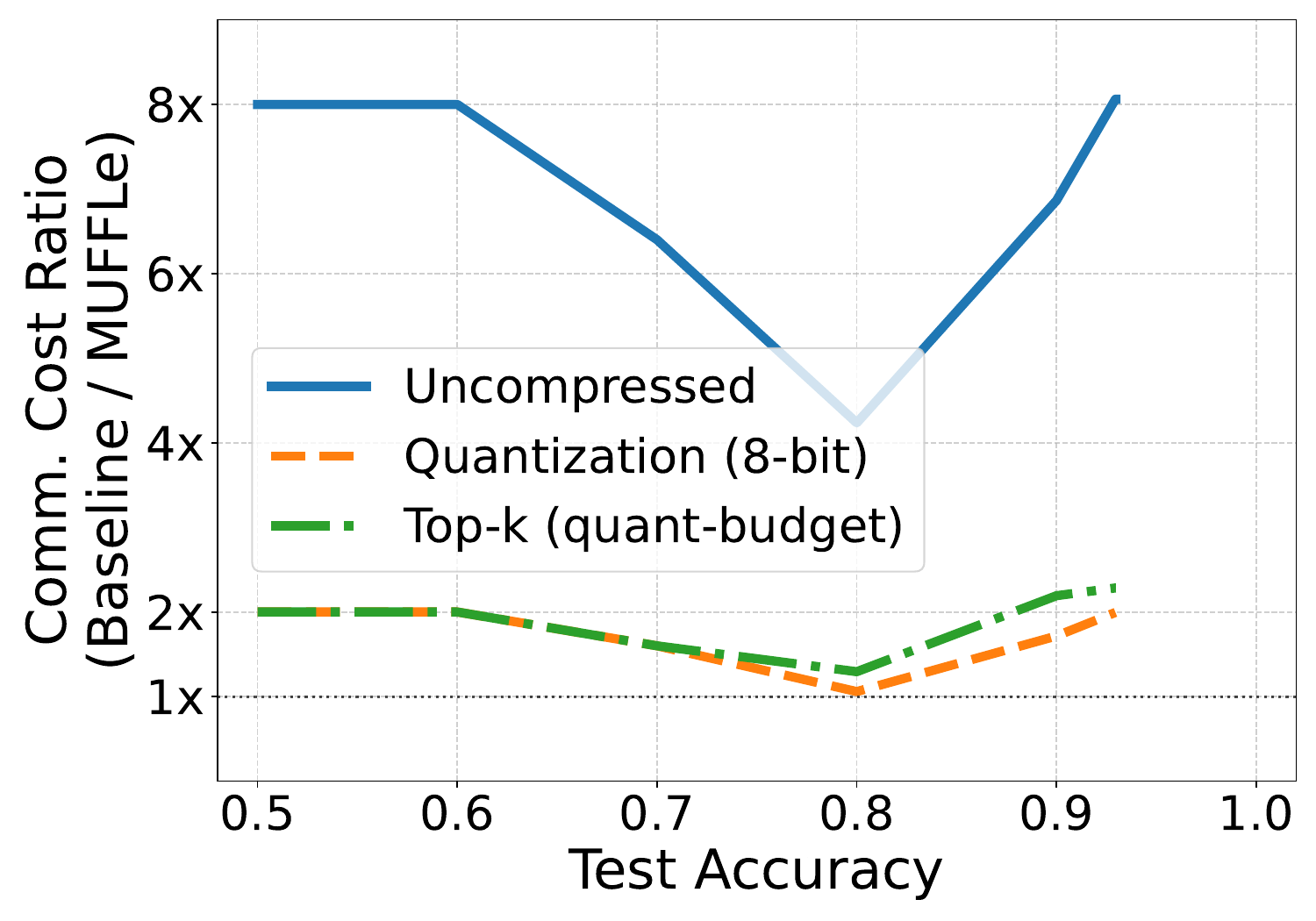}
	\caption{Gains of MUFFLe.}
	\label{fig:cost_reduction}
\end{subfigure}
\caption{Comparison of communication efficiency on MNIST.}
\label{fig:experiment_results}
\end{figure}

Table~\ref{table:mnist_results} summarizes the communication cost required by each method to reach the target accuracy of $92.93\%$, defined as the test accuracy achieved by the uncompressed baseline at 200 rounds.
As shown in Table~I, MUFFLe reaches this target with a cumulative communication cost of 38~MB, compared with 75~MB for 8-bit quantization, 86~MB for Top-$k$ sparsification, and 310~MB for uncompressed FedAvg.
This corresponds to a $8.3$x reduction relative to uncompressed transmission.
Fig.~\ref{fig:acc_vs_comm_cost} further illustrates the communication--accuracy trade-off.
MUFFLe achieves the target accuracy with the lowest cumulative uplink communication among all evaluated methods, indicating that exploiting repeated patterns can substantially reduce transmission cost.
Fig.~\ref{fig:cost_reduction} shows the communication gain of MUFFLe over the baselines across different accuracy levels.
The gain remains substantial throughout and is largest near the target accuracy, reaching up to $8.3\times$ over uncompressed and more than $2\times$ over quantization and Top-$k$.

\begin{table}[h]
\caption{MNIST Results: Communication vs. Accuracy}
\centering
\begin{tabular}{@{}lcccc@{}}
\toprule
Method & Acc. (\%) & Round & Comm. Cost (MB) & Gain \\ \bottomrule
Uncompressed & $92.93$ & $200$ & $310$ & $1\times$ \\
Quantization & $92.93$ & $193$ & $75$ & $4.1\times$ \\
Top-k & $92.93$ & $221$ & $86$ & $3.6\times$ \\
\textbf{MUFFLe} & $92.93$ & $195$ & $\textbf{38}$ & $\textbf{8.3}$$\times$ \\ \bottomrule
\end{tabular}
\label{table:mnist_results}
\end{table}

\section{Conclusion and Future Work}
This WiP paper presented MUFFLe, a communication-efficient update compression scheme for FedAvg.
Results on IID MNIST show that MUFFLe achieves the target accuracy with substantially lower cumulative uplink communication than uncompressed FedAvg, 8-bit quantization, and Top-$k$ sparsification.
These preliminary results demonstrate the feasibility of MUFFLe.
Future work will extend the evaluation to non-IID data, larger models, and error-feedback mechanisms.

\bibliographystyle{IEEEtran}
\bibliography{ref}

\end{document}